\documentclass[10pt,twocolumn,letterpaper]{article}

\RequirePackage{snapshot}

\usepackage{cvpr}
\usepackage{times}
\usepackage{epsfig}
\usepackage{graphicx}
\usepackage{amsmath}
\usepackage{amssymb}
\usepackage{comment}

\usepackage[pagebackref=true,breaklinks=true,letterpaper=true,colorlinks,bookmarks=false]{hyperref}

\cvprfinalcopy 

\def\cvprPaperID{****} 

\ifcvprfinal\fi

\newcommand{\datasetname}{MIT Driving Scene Segmentation}

\usepackage{authblk}

\newcommand\ver[1]{\small\rotatebox{90}{#1}}
\usepackage{booktabs}
\usepackage{subcaption}

\begin{document}

\title{Value of Temporal Dynamics Information in Driving Scene Segmentation}


\author[1]{Li Ding}
\author[1]{Jack Terwilliger}
\author[2]{Rini Sherony}
\author[1]{Bryan Reimer}
\author[1]{Lex Fridman\thanks{Corresponding author: \texttt{fridman@mit.edu}}}

\affil[1]{Massachusetts Institute of Technology (MIT)}
\affil[2]{Collaborative Safety Research Center, Toyota Motor North America}

\maketitle

\begin{abstract}
  Semantic scene segmentation has primarily been addressed by forming representations of single images both with
  supervised and unsupervised methods. The problem of semantic segmentation in dynamic scenes has begun to recently
  receive attention with video object segmentation approaches. What is not known is how much extra information the
  temporal dynamics of the visual scene carries that is complimentary to the information available in the individual
  frames of the video. There is evidence that the human visual system can effectively perceive the scene from temporal
  dynamics information of the scene's changing visual characteristics without relying on the visual characteristics of
  individual snapshots themselves. Our work takes steps to explore whether machine perception can exhibit similar
  properties by combining appearance-based representations and temporal dynamics representations in a joint-learning
  problem that reveals the contribution of each toward successful dynamic scene segmentation. Additionally, we provide
  the \datasetname{} dataset, which is a large-scale full driving scene segmentation dataset, densely annotated for
  every pixel and every one of 5,000 video frames. This dataset is intended to help further the exploration of the value
  of temporal dynamics information for semantic segmentation in video.
\end{abstract}

\section{Introduction}

Forming appearance-based representations of still images with convolutional neural networks (ConvNets) has been
successfully used in object classification, detection, and segmentation tasks \cite{donahue2014decaf, ren2015faster,
  chen2017rethinking}. These representations can be formed via supervised, semi-supervised, or unsupervised methods
\cite{pathak2017learning}. The primary source of information in these approaches is the visual characteristics of a
single image. In contrast, there is evidence that the human visual system largely relies, in real-world
perception tasks, on temporal dynamics information \cite{pan2013eye}, not merely as pre-processing support for object
segmentation but as the primary source of information used for dynamic scene understanding.

Put another way, understanding a static scene in a single image may be a fundamentally different task than understanding
a dynamic scene in video. From this distinction, a number of efforts focused on video object
segmentation have recently emerged \cite{caelles2017one,pont20172017,caelles20182018}. What is not well understood is the degree to which
temporal dynamics of the visual scene in video can contribute to the video scene segmentation task, and consequently the
dynamic scene understanding task.

Our work helps explore the value of temporal dynamics in driving scene segmentation by formulating the appearance-based
and temporal-based as a joint learning problem that reveals the importance of each component for the effective
segmentation of various parts of the driving scene. In addition, we provide the \datasetname{} dataset, which is a
large-scale full driving scene segmentation dataset, densely annotated for every pixel and every one of 5,000 video
frames. The purpose of this dataset is to allow for exploration of the value of temporal dynamics information for full
scene segmentation in dynamic, real-world operating environments.

\section{Related Work}

Semantic segmentation is the most fine-grained form of two-dimensional image region classification (in contrast to image
classification and object detection), and is currently considered to be the frontier of open challenges in computer vision
that seek to interpret visual information. We consider the work on semantic segmentation in still images and in video
separately.

\subsection{Semantic Segmentation in Still Images}

The task of semantic segmentation involves assigning each pixel in the image a label. For object segmentation, a
distinct between foreground and background objects is drawn. For full scene segmentation, both foreground and background
objects must be classified at the level of a pixel
\cite{everingham2015pascal,mottaghi2014role,Cordts2016Cityscapes,zhou2017scene, caesar2016cocostuff}. Over the past five
years, several key adjustments to ConvNet-based architectures have been made to improve segmentation accuracy. First,
dilated convolution (also known as atrous convolution) have been added to address the reduction of resolution due to
pooling or convolution striding while still being able to learn increasingly abstract feature representations
\cite{chen2014semantic, yu2015multi, chen2016deeplab, chen2017rethinking}. Second, several methods have been proposed to
deal with the existence of objects at multiples scales, including (1) image pyramids that deal with the problem at the
image level \cite{farabet2013learning, eigen2015predicting, pinheiro2014recurrent,
  lin2016efficient,chen2016attention,chen2016deeplab}, (2) encoder-decoder structure which deals with the problem at the
multi-scale feature level
\cite{badrinarayanan2015segnet,ronneberger2015u,pohlen2017fullresolution, peng2017large}, (3) cascading extra modules that deals with the problem by capturing long-range context
\cite{krahenbuhl2011efficient, chen2014semantic, zheng2015conditional, lin2016efficient, schwing2015fully,
  liu2015semantic, yu2015multi}, and (4) spatial pyramid pooling that deals with the problem by using filters and
pooling operations of various rates and sizes \cite{chen2016deeplab, zhao2016pyramid}.

Overall, progress in semantic segmentation of still images has continued \cite{chen2017rethinking}, and it is possible
that eventually any approach to semantic segmentation in video will eventually completely disregard temporal dynamics of
the scene, as it has for the state-of-the-art tracking by detection approaches. However, this possible eventuality is
far from guaranteed, and is currently one of the open problems of computer vision: how valuable is temporal dynamics
information for scene understanding in video? Our work seeks to take steps toward answering this question.

\subsection{Semantic Segmentation in Video}

Most of the work in semantic segmentation has been on the problem of \emph{video object segmentation} where a
distinction is drawn between foreground and background objects, and the task is focused on temporal propagation of
foreground object segmentation. A wide variety of approaches have been proposed for this task. The first set of
approaches groups pixels spatiotemporally based on motion features computed along individual point trajectories
\cite{brox2010object, ochs2011object, keuper2015motion, fragkiadaki2012video}. These approaches rely on successful
feature matching in the temporal domain, and fail when such matching is intermittently erroneous. The second set of
approaches formulates segmentation as a foreground-background classification task, detecting regions that correspond to
foreground objects and matching the resulting appearance models with other information such as salience maps, shape
estimates, and pairwise constraints \cite{papazoglou2013fast,lee2011key,wang2015saliency}. The third set of approaches
incorporate the classification approach with a memory module for propagating region estimates in time
\cite{tokmakov2017learning,xu2018youtube}. The latter set of approaches begin to incorporate temporal dynamics
information into the learning problem, and due to their success, motivate the method, observation, and dataset proposed
in our work. The primary benchmark dataset used for semantic segmentation in video to date is DAVIS
\cite{pont20172017,caelles20182018}. Our dataset differs in three ways: (1) it provides full scene segmentation not just
foreground object segmentation, (2) it includes only driving scenes from the perspective of the ego-vehicle, and (3) it
is densely annotated in time for long period at 30 fps.


\begin{figure*}
	\begin{subfigure}{.24\textwidth}
		\centering
		\includegraphics[width=.99\linewidth]{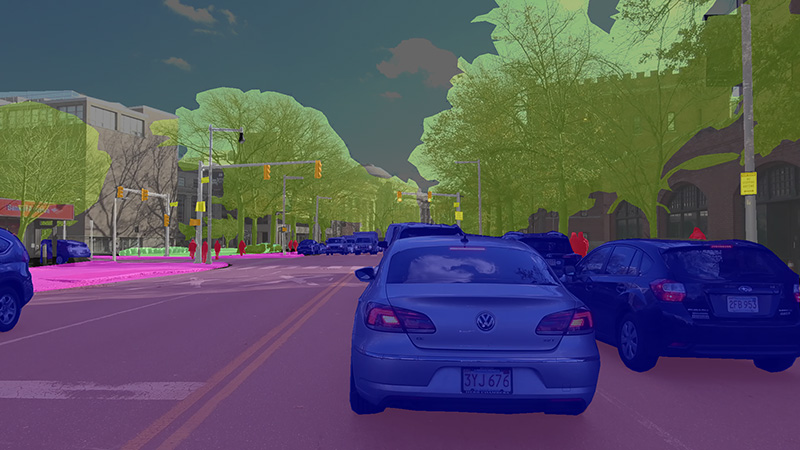}
	\end{subfigure}
	\begin{subfigure}{.24\textwidth}
		\centering
		\includegraphics[width=.99\linewidth]{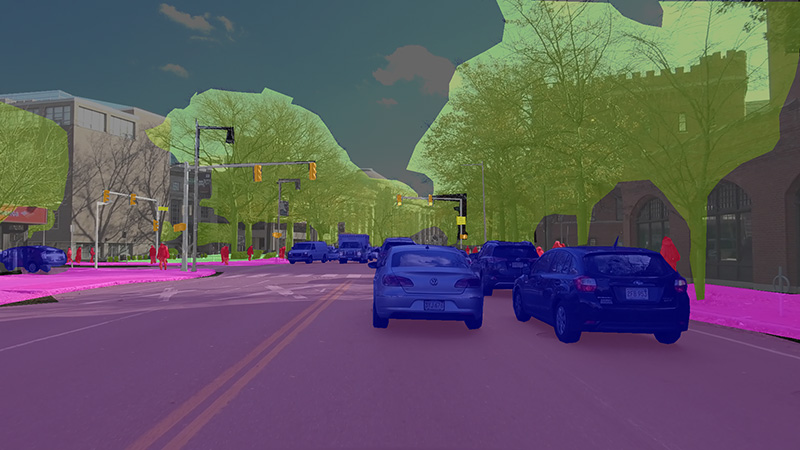}
	\end{subfigure}
	\begin{subfigure}{.24\textwidth}
		\centering
		\includegraphics[width=.99\linewidth]{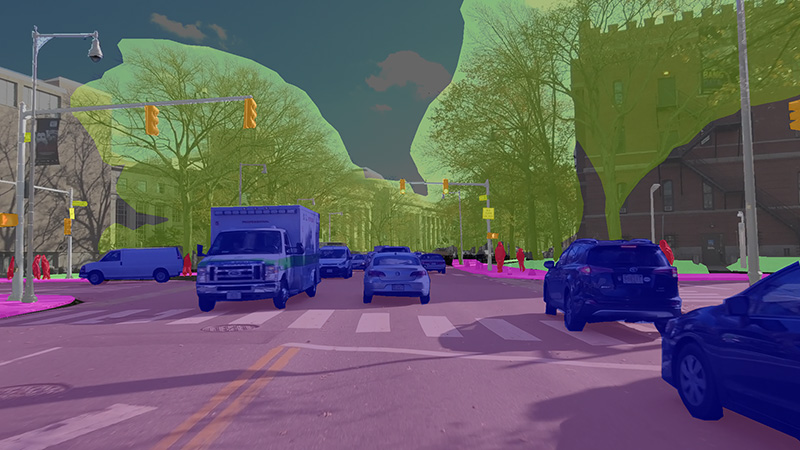}
	\end{subfigure}
	\begin{subfigure}{.24\textwidth}
		\centering
		\includegraphics[width=.99\linewidth]{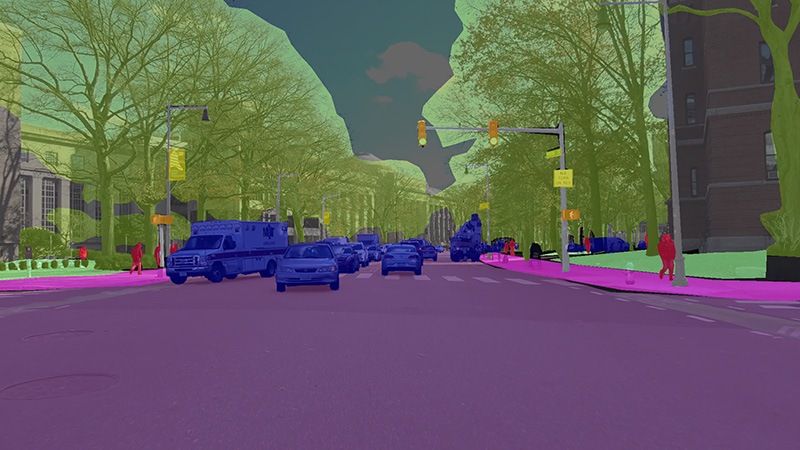}
	\end{subfigure}
	\par\smallskip
	\begin{subfigure}{.24\textwidth}
		\centering
		\includegraphics[width=.99\linewidth]{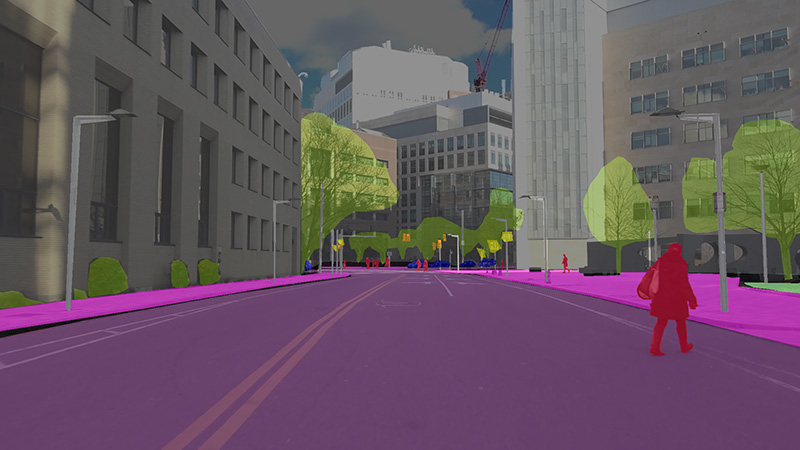}
	\end{subfigure}
	\begin{subfigure}{.24\textwidth}
		\centering
		\includegraphics[width=.99\linewidth]{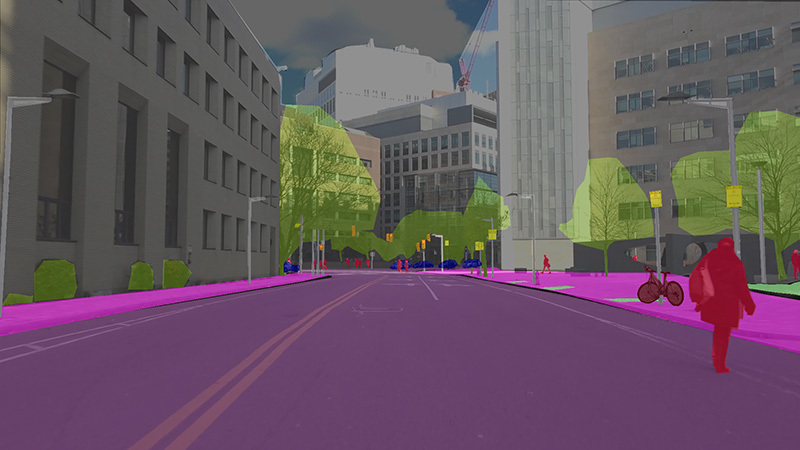}
	\end{subfigure}
	\begin{subfigure}{.24\textwidth}
		\centering
		\includegraphics[width=.99\linewidth]{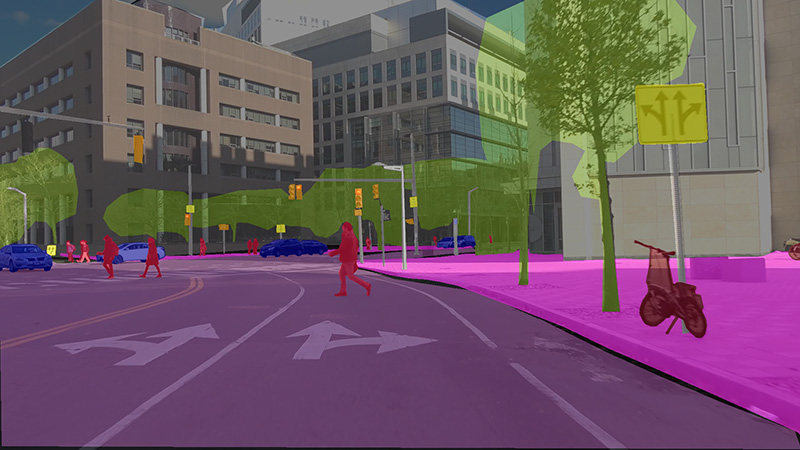}
	\end{subfigure}
	\begin{subfigure}{.24\textwidth}
		\centering
		\includegraphics[width=.99\linewidth]{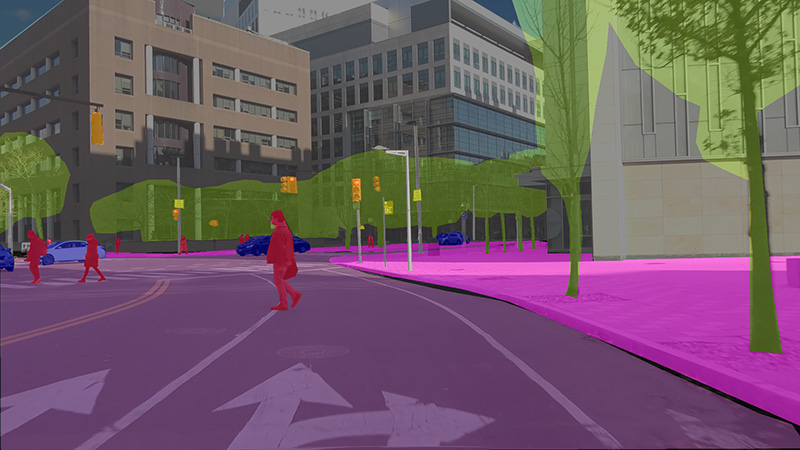}
	\end{subfigure}
	\par\smallskip
	\begin{subfigure}{.24\textwidth}
		\centering
		\includegraphics[width=.99\linewidth]{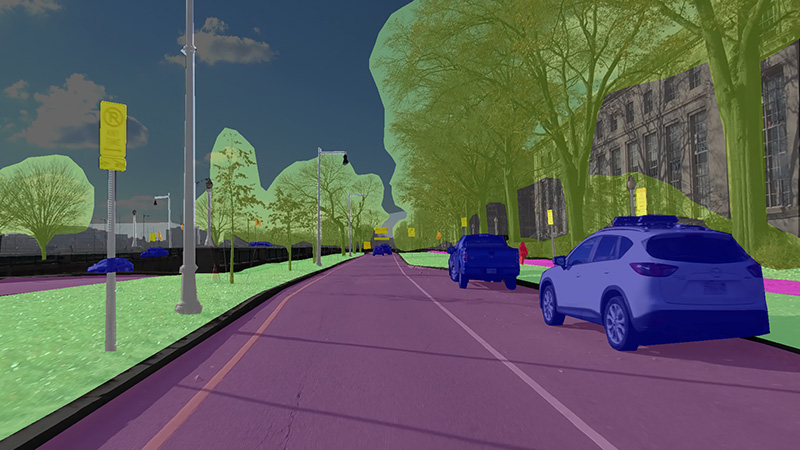}
	\end{subfigure}
	\begin{subfigure}{.24\textwidth}
		\centering
		\includegraphics[width=.99\linewidth]{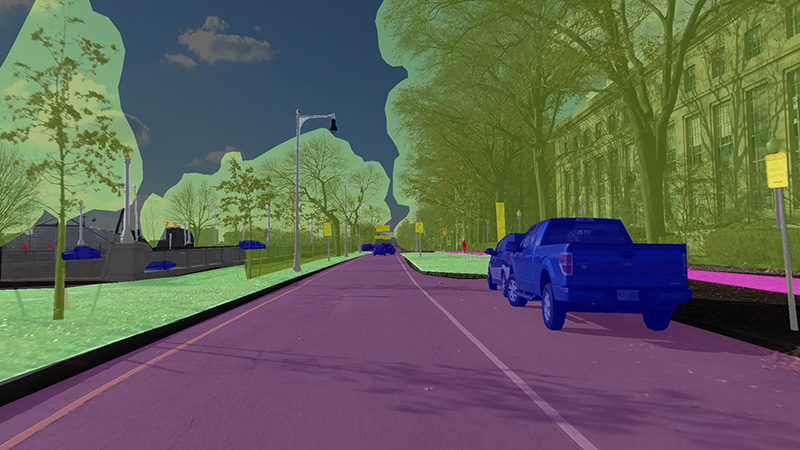}
	\end{subfigure}
	\begin{subfigure}{.24\textwidth}
		\centering
		\includegraphics[width=.99\linewidth]{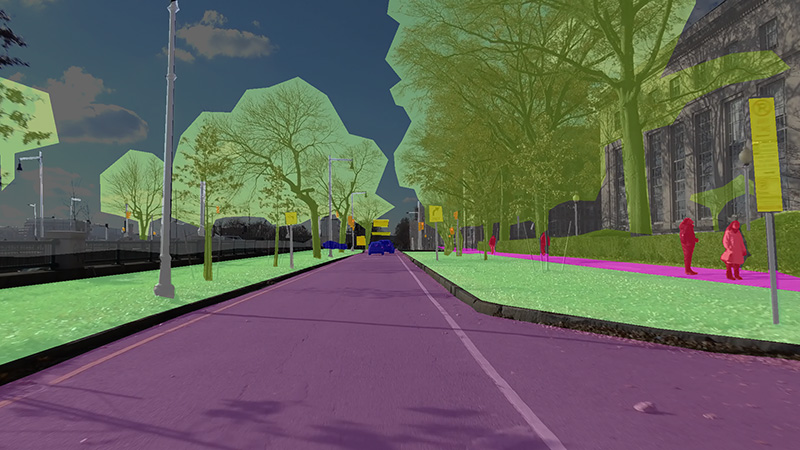}
	\end{subfigure}
	\begin{subfigure}{.24\textwidth}
		\centering
		\includegraphics[width=.99\linewidth]{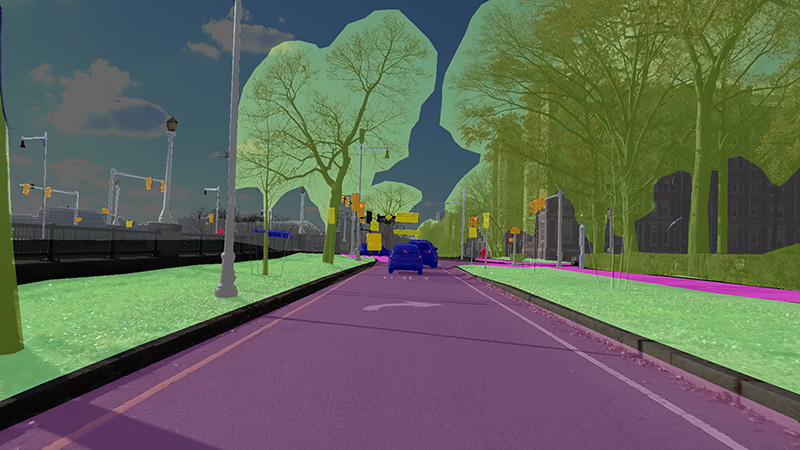}
	\end{subfigure}
	\par\smallskip
	\begin{subfigure}{.24\textwidth}
		\centering
		\includegraphics[width=.99\linewidth]{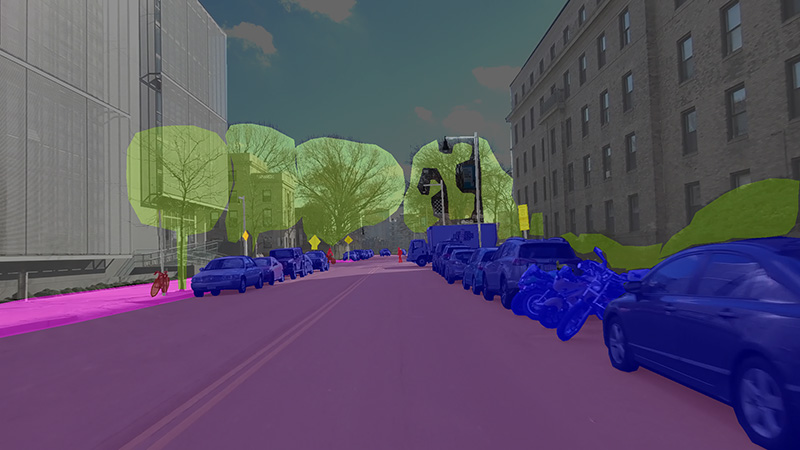}
	\end{subfigure}
	\begin{subfigure}{.24\textwidth}
		\centering
		\includegraphics[width=.99\linewidth]{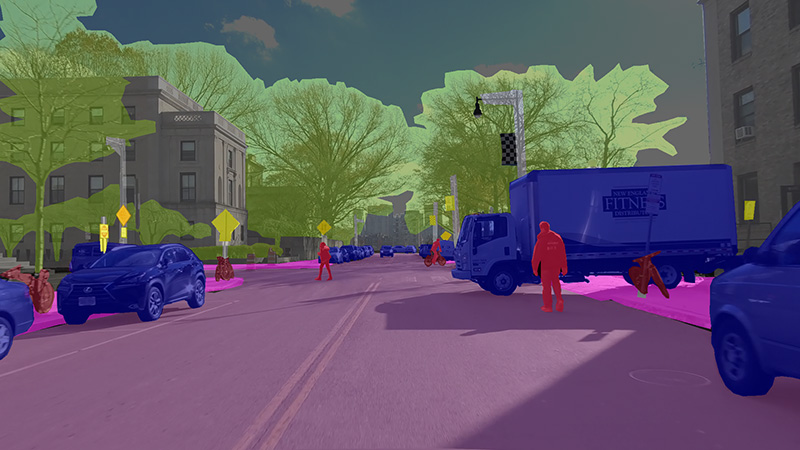}
	\end{subfigure}
	\begin{subfigure}{.24\textwidth}
		\centering
		\includegraphics[width=.99\linewidth]{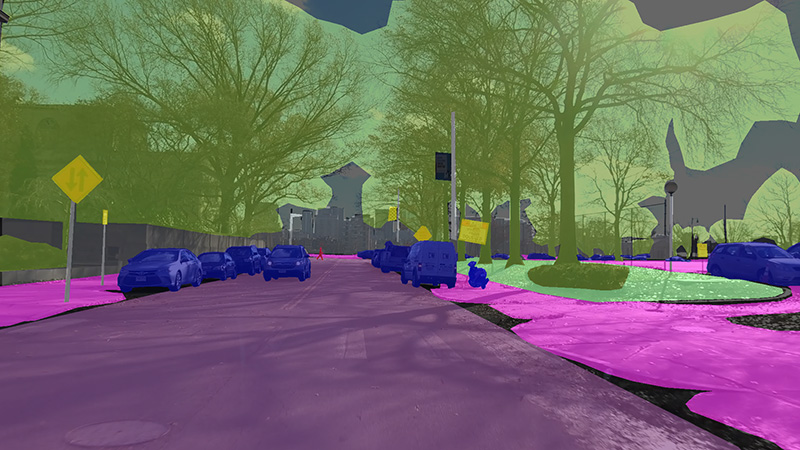}
	\end{subfigure}
	\begin{subfigure}{.24\textwidth}
		\centering
		\includegraphics[width=.99\linewidth]{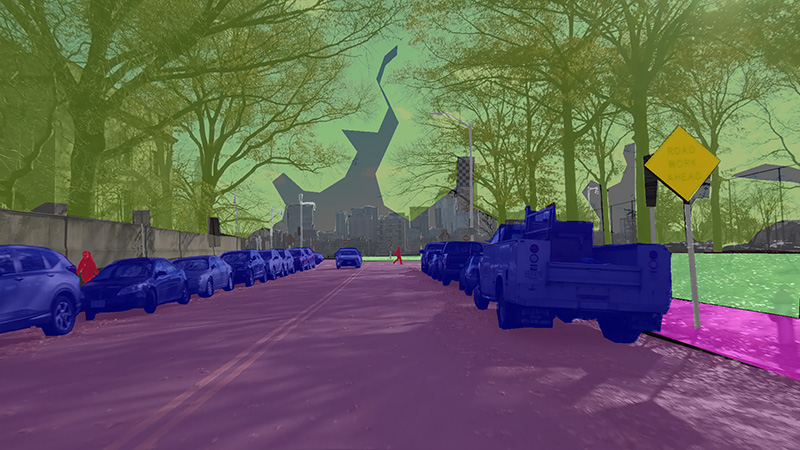}
	\end{subfigure}
	
	\caption{Examples from the proposed  \datasetname{} dataset. Annotations are overlayed on frames.}
	\label{fig:dataset}
\end{figure*}

\section{Method}

Deep neural networks being pre-trained on large-scale datasets show a better capability of generalization after fine-tuning even without using any temporal information~\cite{caelles2017one, maninis2018video}. In addition, \cite{carreira2017quo} also shows that there is a considerable benefit in pre-training on large video dataset other than only still images. However, in the context of video scene parsing, it is too costly to obtain a densely annotated video dataset at large-scale. In order to address the advantage of both pre-trained networks on still images and pre-trained networks on videos, our method focus on combining the appearance network (pre-trained on still images) and the memory network (pre-trained on videos) in a meaningful way, while under the assumption that the appearance network itself already well-performs.

\subsection{Generalizing Semantic Visual Memory}

Recent work successfully use convolutional recurrent networks, \eg Conv-LSTM~\cite{xingjian2015convolutional} and Conv-GRU~\cite{tokmakov2017learning}, to address spatiotemporal sequence modeling problem. However, due to the requirement of large-scale data in training deep neural networks, such models usually either have to hold a insufficient small number of hidden states or suffer the lack of ability of generalization. In this case, we build a memory network using Conv-LSTM taking the deep semantic feature map from a pre-trained appearance network as input, and pre-train it again on a large-scale video segmentation dataset. Specifically, we adopt the DeeplabV3 network~\cite{chen2017rethinking} as the appearance network, and use a Conv-LSTM with 256 hidden units as the memory network. By freezing the weights of the appearance network when training the memory network, we avoid the risk of losing the ability of generalization when feeding highly redundant dense video frames.

\subsection{Prediction Refinement with Confidence Gates}

From the sense of human perception, it is intuitive that some parts of the scene are easier to perceive statically, others with motion instead. However, it is unclear how this should be managed and distributed throughout the whole image. We design the confidence gate of prediction update, inspired by the way how LSTM updates its cell states.

For each frame, given the prioritary class prediction from appearance network $logits_{appr}$, and the prediction from memory network $logits_{mem}$, which is obtained from a 1x1 Conv layer after Conv-LSTM, the final prediction is calculated as 
\[
logits = \sigma_{appr} \cdot logits_{appr} + \sigma_{mem} \cdot logits_{mem}
\]
where $\sigma_{appr}$ and $\sigma_{mem}$ are two spatial sigmoid gates, calculated from two 1x1 Conv layers taking the concatenation of appearance network feature and Conv-LSTM output as input. The gates control how to distribute the weights, \ie confidence on the results provided by both parts of the system. The appearance network except for the last layer is frozen during training due to the risk of over-fitting, although the system is end-to-end trainable.

\subsection{Training Process} \label{sec:training_process}

The goal for our training process is to combine state-of-the-art methods developed on both still images and videos, while at the same time preserve the natural advantage of both. The target testing case is doing semantic segmentation on video frames, while the training can be done either on still images only or with video frames. However, the memory network requires sequences of video frames for training, which should have at least one frame annotated. 

In this case, we first adopt the DeeplabV3 pretrained on still images as feature extractor, and train the memory network on a video dataset. Then we fine-tune the memory network on the target dataset with multiple frames but only calculate the loss from the last frame, where the ground truth segmentation exists. Finally, we fine-tune the whole system with confidence gates on the output from both appearance network and memory network, while not changing the weights in feature extractor layers.

To further explain our idea in the above design, we consider the scenario that the appearance network is good enough, but there will be some edge cases that can not be dealt with using only one still image, such as occlusion, cut on image margin, motion blur, \etc So the memory network is to help with those cases, and the confidence gates are used to control the merging of information.

\section{Dataset}

Since our target problem is video scene parsing in the driving context, we group current datasets into three categories: 1) Pixel-wise annotation of still images, \eg Mapillary Vistas~\cite{neuhold2017mapillary}; 2) Pixel-wise annotation of coarse video frames, \eg Cityscapes~\cite{Cordts2016Cityscapes}, BDD~\cite{yu2018bdd100k}; 3) Pixel-wise annotation of dense video frames, \eg the \datasetname{} dataset in this work. Category 1 is the easiest to obtain and get larger variability in different scenes. Category 3 usually lack variability, but is the most suitable to do temporal modeling.

\begin{figure*}
	\centering
	\includegraphics[width=.9\textwidth]{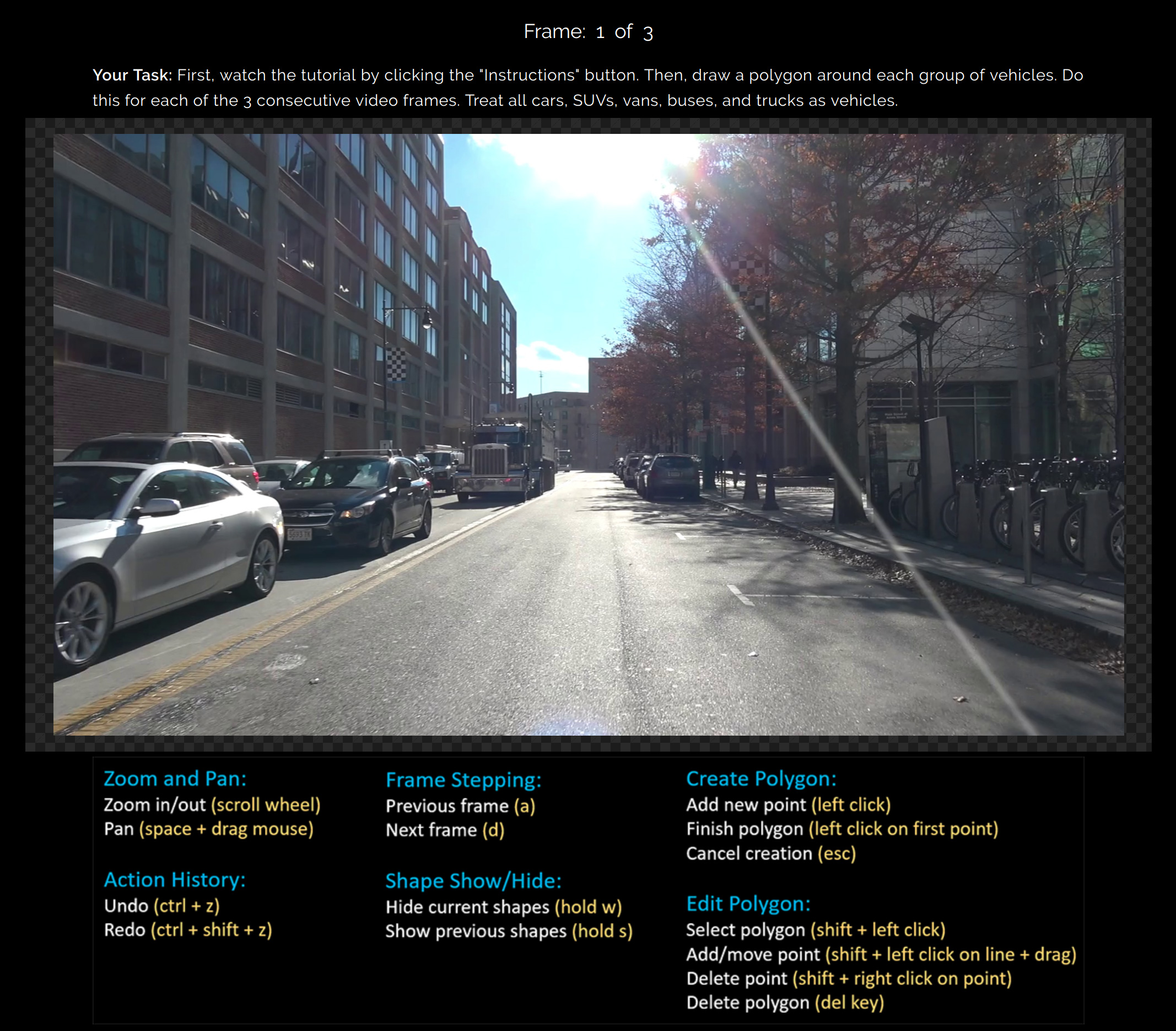}
	\caption{Front-end of our annotation tool.}
	\label{fig:tool}
\end{figure*}

\subsection{Dataset Selection}
There are many large-scale datasets with semantic pixel-wise annotations, \eg Pascal VOC~\cite{Everingham10}, MS COCO~\cite{lin2014microsoft}, ADE20k~\cite{zhou2017scene}, but more about natural scene/object. 
In the driving context, there are several well-developed datasets with dense semantic annotations, \eg CamVid~\cite{brostow2009semantic}, KITTI~\cite{geiger2013vision}, Cityscapes~\cite{Cordts2016Cityscapes}, Mapillary Vistas~\cite{neuhold2017mapillary}, and recently BDD~\cite{yu2018bdd100k}. 
Among the above datasets, Mapillary Vistas contains 25k images with fine annotation, which is the largest value, but the images are all still images, \ie without temporal connection between each other. 
Both Cityscapes and BDD choose one frame from each short video clip to cast fine annotation.
It is a common approach to make the dataset such way in order to mostly capture the variability of scenes with the least amount of budget. 
However, this phenomenon also leads current research to focus on single-frame algorithms, ignoring the rich temporal information contained between consecutive frames. 

In this case, we collect and annotate a novel dataset, which has over 10k frames with fine annotation, from a single,
untrimmed video of the front driving scene at 30 fps.  The similar idea can be seen in CamVid, which also features
annotated video frames, but only at a low frequency (1 fps) and a small scale (less than 500 frames totally).  Out of
driving scene domain, DAVIS~\cite{perazzi2016benchmark} has densely annotated pixel-wise semantic object annotations for
trimmed video clips, each at around 60 frames. Similarly, SegTrack~\cite{TsaiBMVC10} and SegTrack v2~\cite{FliICCV2013}
also feature pixel-wise video object annotation.

In order to make our approach comparable with other recent work, we choose Cityscapes as our main source of data in experiments. 
Cityscapes is the largest dataset focused on urban street views, which contains 5k finely-annotated and 20k coarsely-annotated images. 
The \datasetname{} dataset is mainly used for pre-training in this work, which is described in detail in Sec.~\ref{sec:dataset}

\subsection{\datasetname{} Dataset} \label{sec:dataset}

The two main purposes of the development of \datasetname{} dataset are: 
1) experiment and develop the full-scene annotation system that is scalable with a large pool of workers, \eg on Amazon Mechanical Turk (MTurk); 
2) create an open-source densely-annotated video driving scene dataset that can help with future research in various fields, \eg spatiotemporal scene perception, 
predictive modeling, semi-automatic annotation process development. 
In this case, we collect a long, untrimmed video (6:35, 11869 frames) at 1080P (1920x1080), 30 fps, which is a single daytime driving trip, and annotate it with fine, 
per-frame, pixel-wise semantic labels. Examples from the dataset are shown in Fig.~\ref{fig:dataset}.
Some other kinds of driving-related metadata such as IMU, GPS are also available.

Existing driving scene datasets rely on hiring a very small group of professional annotators to do full frame annotation, which usually takes around 1.5 hours per 
image.~\cite{Cordts2016Cityscapes, neuhold2017mapillary}
This is reasonable because the pixel-wise annotation is a task of high-complexity that reach the limits of human perception and motor function without specific training. 
However, in order to have scalability and flexibility to annotate in potentially much larger scale, we develop the annotation tool that is web-based and understandable. 
We deploy our tool on MTurk, which contains a large pool of professional and non-professional workers. 

\subsubsection{Annotation Tool}

To support fine and quick annotations, we develop a web-based annotation tool following the common polygon annotator~\cite{russell2008labelme} design with the 
implementation of techniques such as zoom in/out and keyboard shortcuts. A screenshot of our annotation tool is shown in Fig~\ref{fig:tool}.
To further promote the accuracy and efficiency of annotation processes, we also address several problems during the process and design specific improvements.

\textbf{Task complexity.} \hspace{1em}
We found that annotators had difficulty when asked to annotate an entire scene at once, which involved keeping in mind many object classes and keep working for a non-flexible long time. 
In response, we divided the task of annotating an entire scene into multiple subtasks, in which an annotator is responsible for annotating only 1 object class. 
We found that this largely reduces classification errors, improves the quality of our annotations, and reduces the time required to fully annotate each scene.

\textbf{Limits of human perception.} \hspace{1em}
We found many misclassification errors are due to the difficulty of recognizing objects in still scenes, but that these errors appeared obviously incorrect in the video. 
In response, we designed the tool such that an annotator could step through consecutive frames quickly with keyboard shortcuts. 
The motion perceived when stepping through frames, reduce classification errors.

\begin{table*}[]
	\begin{center}
		\footnotesize
		\begin{tabular}{c@{\hspace{2mm}}|@{\hspace{2mm}}*{19}{@{\hspace{2mm}}c}@{\hspace{2mm}}|@{\hspace{2mm}}c}
			\small{Method}                 & \ver{Road}  & \ver{Sidewalk} &\ver {Building} & \ver {Wall } & \ver {Fence} & \ver {Pole } & \ver {Traffic Light} & \ver {Traffic Sign} & \ver {
				Vegetation} & \ver {Terrain} & \ver{Sky  } & \ver{Person} & \ver{Rider} & \ver {Car  } & \ver {Truck} & \ver {
				Bus  } & \ver {Train} & \ver {Motorcycle} & \ver {Bicycle} & \ver {Mean IoU} \\ \midrule
			*DeeplabV3~\cite{chen2017rethinking}              & 97.5 & 81.0    & 90.3    & 38.4 & 53.8 & 50.8 & 61.4         & 71.3        & 91.0      & 58.9   & 93.0 & 76.3  & 53.2 & 93.2 & 69.2 
			& 
			75.7 & 
			63.7 & 55.2      & 72.6   & 70.9         \\
			Ours (Cityscapes only) & 97.6 & 81.3    & 90.5    & 41.0 & 55.0 & 50.6 & 61.6         & 71.0        & 91.0      & 61.5   & 92.9 & 76.1  & 51.8 & 93.2 & 68.9 & 75.6 
			& 62.4 & 54.4      & 72.4   & 71.0         \\
			Ours (full)                  & 97.6 & 81.3    & 90.6    & 44.5 & 55.4 & 51.0 & 61.8         & 71.3        & 91.0      & 62.0   & 92.9 & 76.2  & 52.8 & 93.4 & 70.2 & 77.1 & 64.9 
			& 55.6      & 72.4   &  71.7     
		\end{tabular}
		
		\caption{Quantitative results on Cityscapes dataset. The proposed model achieve better results on stuff classes, \eg sidewalk, wall, fence, terrain. The 
			improvement is mostly contributed by the memory network, since we fix the weights of all the feature extractor layers in the appearance network. (*We use the 
			officially released DeeplabV3 model checkpoint with MobileNetV2~\cite{sandler2018inverted} backbone pre-trained on MS-COCO~\cite{lin2014microsoft}. The 
			same below.)}
		\label{table1:results}
	\end{center}
\end{table*}

\subsubsection{Annotation Process}
The intuition behind our annotation process, is that small simple tasks are preferable to large complex tasks. 
By breaking down the semantic segmentation task into sub-tasks so that each worker is responsible for annotating only part of a scene, the annotations are: 1) easier to validate 2) easier and more efficient to annotate and 3) higher quality.
To accomplish this, we divide the work of annotating the video into tasks of 3 frames in which a worker is asked to draw polygons around only 1 class of object, \eg vehicles. 
Our annotation process involves 4 stages: 1) task creation 2) task distribution 3) annotation validation and 4) the assembly of sub-scene annotations into full-scene annotations.

For stage 1, the creation of tasks, we label which frames contain the classes we are interested in and group the frames into sets of 3.
This stage removes cases where a worker is asked to annotate the presence of a class which is not present in a frame. 
Since this stage only requires labeling frame numbers in which a member of a particular class enters a visual scene and frame numbers in which the last member of a particular class leaves, it is much faster and cheaper than creating a semantic segmentation task for every frame and let annotators find out the class does not exist. 
This approach creates significant time-and-cost-savings especially for rare classes, such as motorcycles in our case. 

For stage 2, the distribution of tasks, we submit our tasks to MTurk and specify additional information which controls how our tasks are distributed:
\begin{itemize}
    \item Reward: This is the amount of money a worker receives for completing our task. We specify different rewards for different classes based on the estimated duration and effort in the annotation.
    \item Qualifications: This allows us to limit the pool of workers who may work on our tasks based on 
1) the worker’s approval rate, or rate of successful annotation, calculated from all a worker’s work on the MTurk platform.
2) the total number of tasks the worker has completed.
3) the qualification task we designed for every new worker taking our task for the first time, which is a test task that can be evaluated with the known ground truth.
\end{itemize}{}

For stage 3, annotation validation, we use both automated and manual processes for assessing the quality of worker annotations. 
In addition to the first qualification task, workers are assigned additional test tasks occasionally, which are indistinguishable from non-ground truth tasks, to check whether they are still following our instructions. 
If the worker’s annotation deviates significantly from the ground truth, they are disqualified from working on our tasks in the future.
The process of comparing worker’s annotations with the ground truth is automated, by calculating the Jaccard distance. 
The threshold score is class dependent since it is easier to score high on less-complex objects like the road than pedestrians.
For our manual validation process, we visually validate that a worker’s annotations are of sufficient quality, using a tool which steps through annotated frames as a video player which allows approving/rejecting work and blocking workers via key presses.

For stage 4, the merging of sub-annotations, we combine the class-level annotations for a given frame into a full-scene annotation. 
For this task, we automatically compose the final full-scene annotation one class at a time. 
Our algorithm first draws the background classes, such as road and sidewalk, and then stationary foreground objects, such as poles
and buildings, and finally dynamic foreground objects such as pedestrians and vehicles.
In order for this to work, we carefully designed the instructions for each class so that they could fit together harmoniously. 
The order in which we draw the classes dictates the instructions. 
When annotating the $i^{th}$ class of n total classes, a worker must annotate the boundaries between objects of class $i$ and classes $j$ where $j >= i$. 
In other words, if we draw the road annotations before vehicle annotations, workers do not need to draw the boundary between road and vehicle when annotating road, since this work will be handled by workers annotating vehicles.

\section{Experimental Results}

We do experiments on Cityscapes and the proposed \datasetname{} dataset, following the training process described in Sec.~\ref{sec:training_process}. The quantitative and qualitative results are both reported on the validation set of Cityscapes.

\subsection{Quantitative Results}

The quantitative results are calculated using the evaluation scripts provided by \cite{Cordts2016Cityscapes}, shown in Table~\ref{table1:results}. For Cityscapes only 
model, we train the memory network from scratch using only sequences of frames and calculate the loss on the last frame with ground truth. Note that our model is 
designed to be causal, not using future information, which is capable to run in real time. 

The two major findings from the quantitative results are: 
1) With the memory network pre-trained on the proposed \datasetname{} dataset, the overall performance gets improved. The improvement is mostly contributed by 
the memory network, since we fix the weights of all the feature extractor layers in the appearance network. In other words, even with a fixed appearance network that 
is already well-trained on still images, we can still find clues to improve it from the temporal domain. 
2) By adding the memory network, the model performs better on stuff classes, \eg sidewalk, wall, fence, terrain, which indicates that these classes are preferred by the 
memory network. We have further exploration on this point described in the next section with qualitative examples.

\subsection{Qualitative Results} \label{Sec:QR}

The purpose of getting qualitative results is to further reveal and help understand the advantage of having a memory network to model spatiotemporal information.

\subsubsection{Border Denoising for Stuff Classes}

\begin{figure*}
	\begin{subfigure}{.33\textwidth}
		\centering
		\includegraphics[width=.99\linewidth]{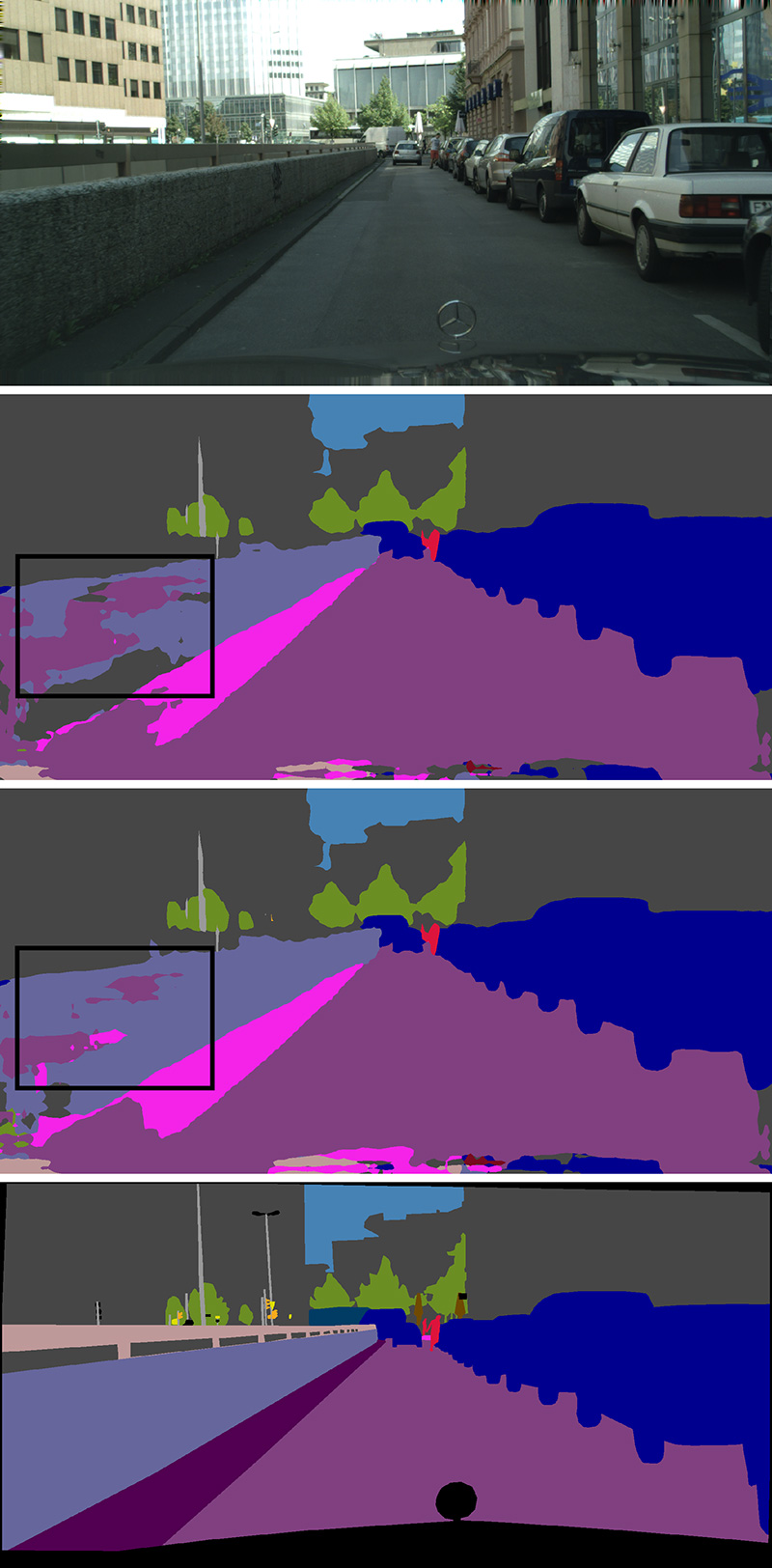}
	\end{subfigure}
	\begin{subfigure}{.33\textwidth}
		\centering
		\includegraphics[width=.99\linewidth]{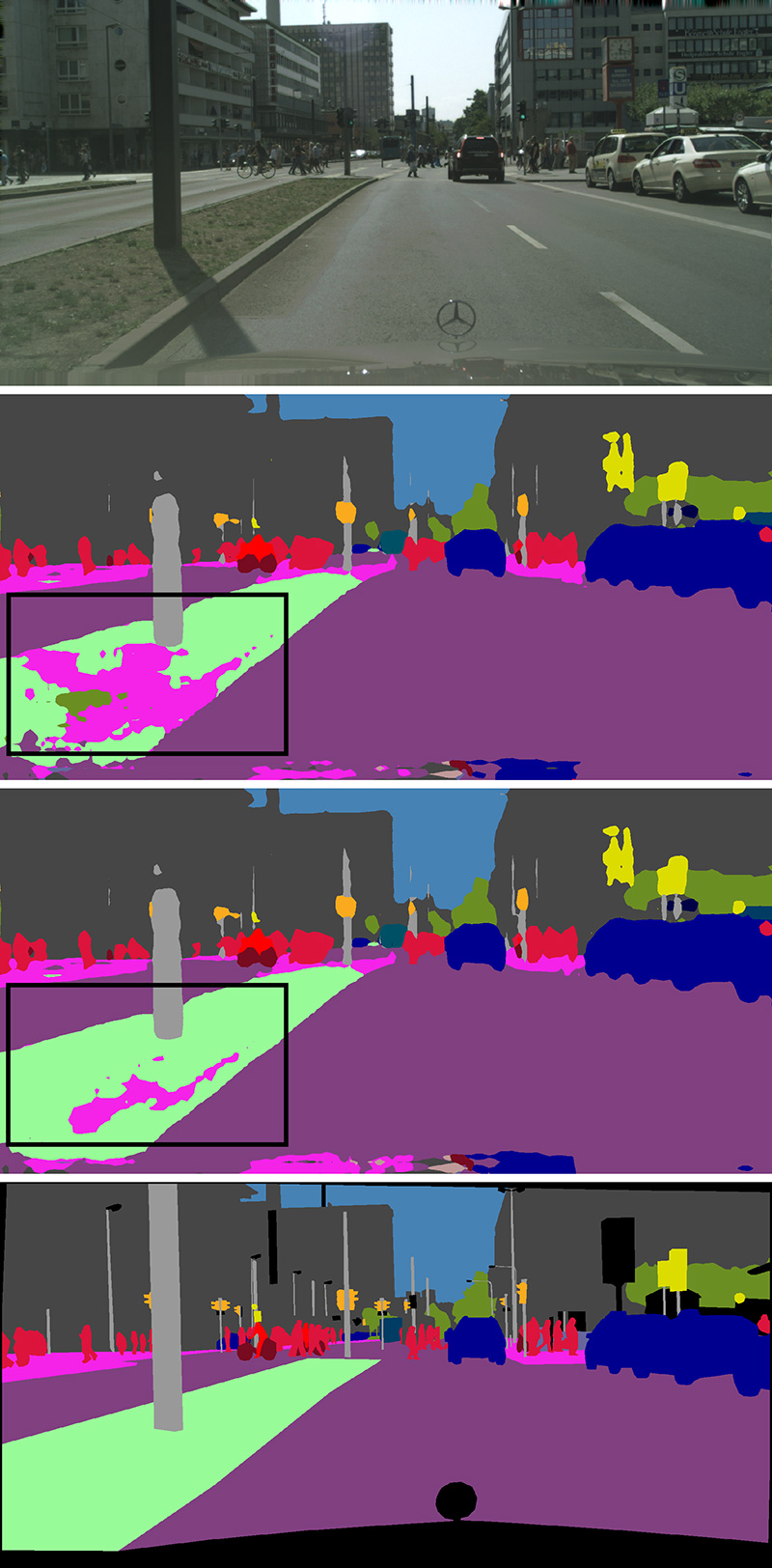}
	\end{subfigure}
	\begin{subfigure}{.33\textwidth}
		\centering
		\includegraphics[width=.99\linewidth]{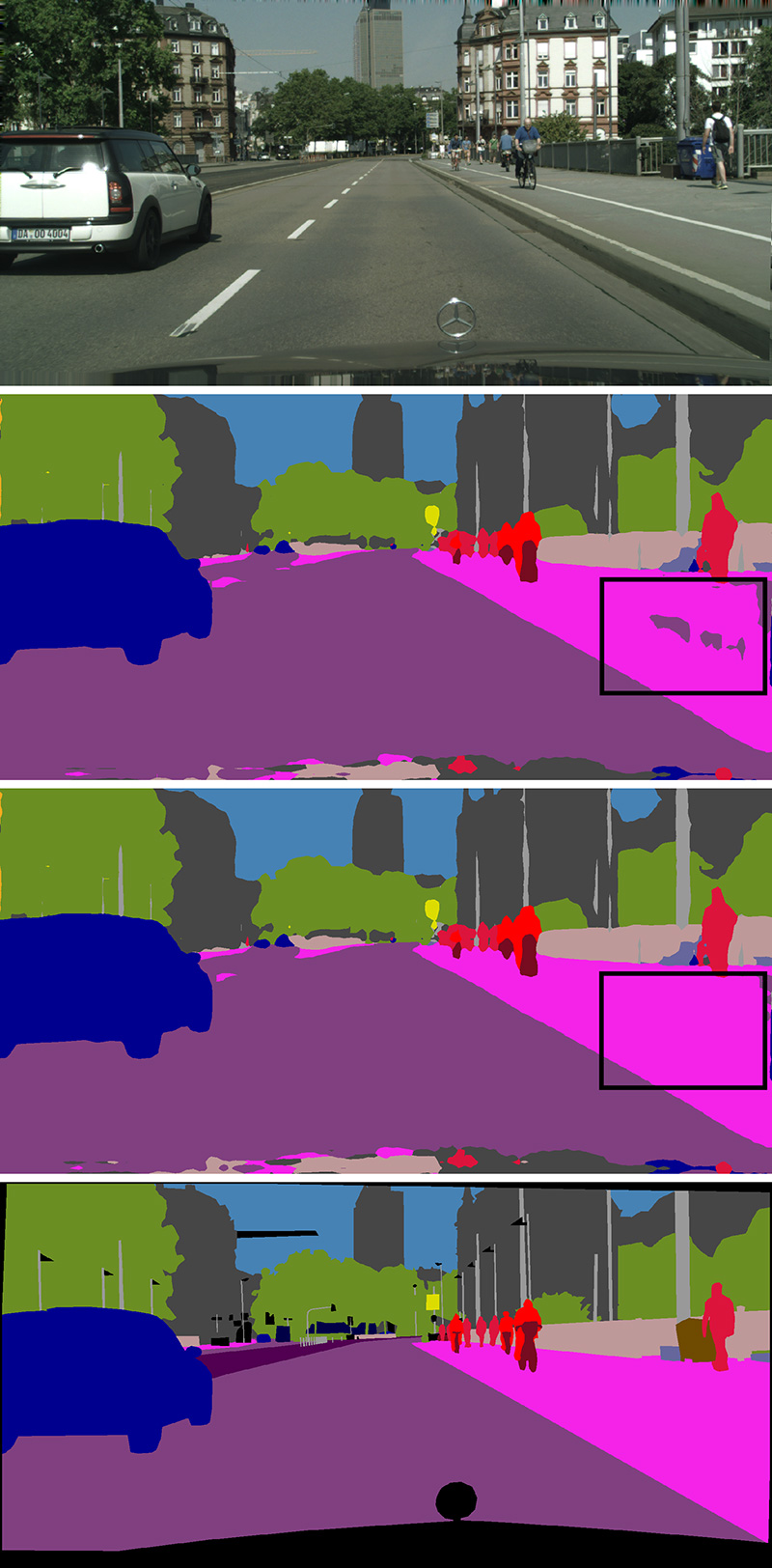}
	\end{subfigure}
	\caption{Visualization of segmentation results. From top to bottom: input image, DeeplabV3 results, our results, ground truth. Highlighted areas show the 
	improvement on border area with stuff classes (from left to right: wall, terrain, sidewalk).}
	\label{fig:border}
\end{figure*}

We visualize some of the cases where there are visible improvement of our results over the baseline DeeplabV3, as shown in Fig.~\ref{fig:border}. The baseline model 
fails to predict some stuff classes on the border area of the image, possibly due to camera effect, motion blur, or lack of context. Our proposed method is able to 
reduce this kind of mistakes by taking consideration of spatiotemporal context with the memory network. This finding aligns with the quantitative results where we 
find most of the improvement of our method lies in the stuff classes. It is important to take this consistency problem seriously in the driving domain, since the driving 
decisions are always made within a very short time, and the image quality is hard to always maintain due to the variability of dynamic driving scene.

\subsubsection{Confidence Gates as Distributed Attention on Spatial v.s. Spatiotemporal Information}

\begin{figure*}
	\begin{subfigure}{.33\textwidth}
		\centering
		\includegraphics[width=.99\linewidth]{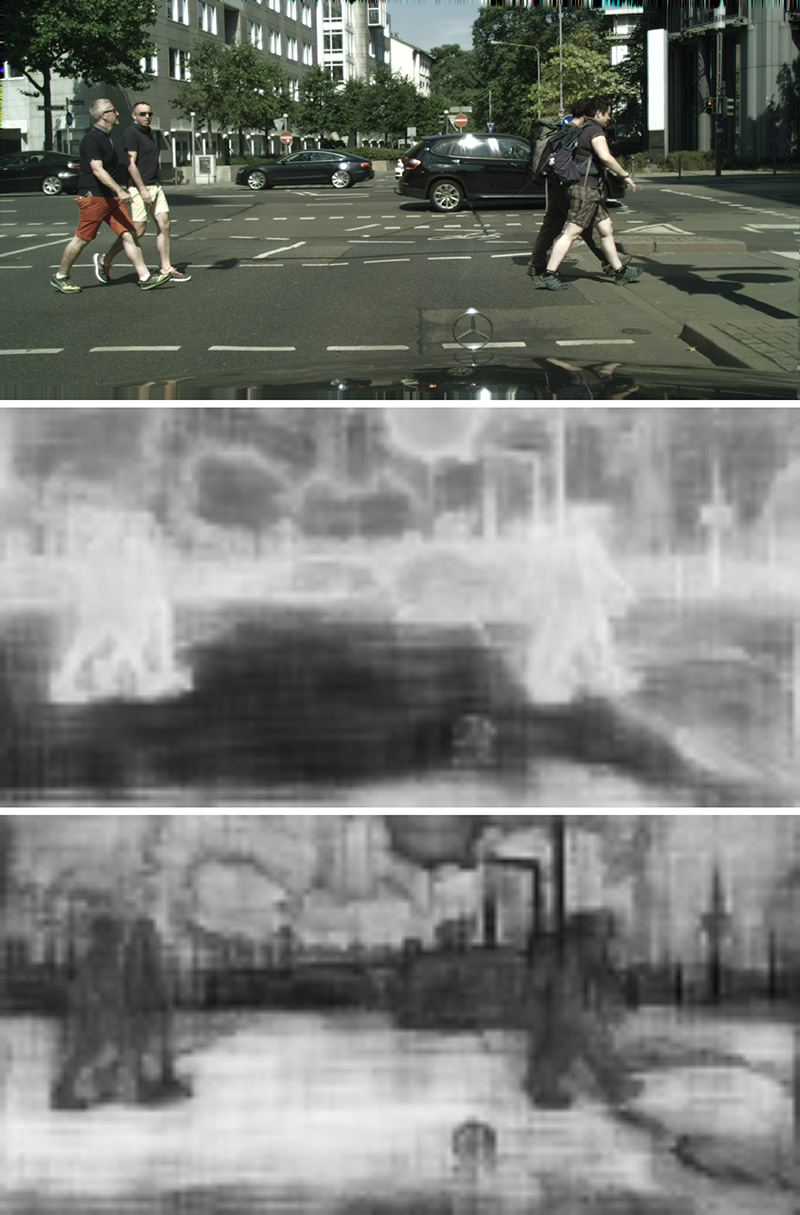}
	\end{subfigure}
	\begin{subfigure}{.33\textwidth}
		\centering
		\includegraphics[width=.99\linewidth]{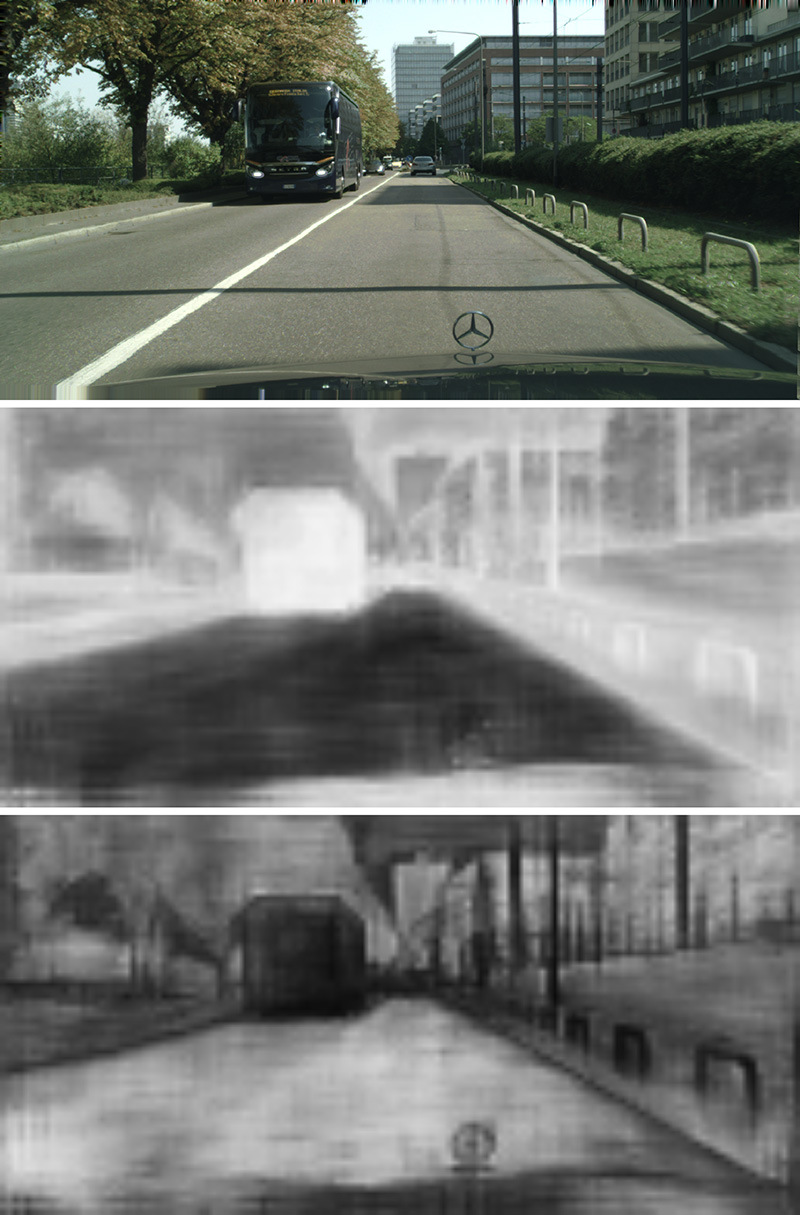}
	\end{subfigure}
	\begin{subfigure}{.33\textwidth}
		\centering
		\includegraphics[width=.99\linewidth]{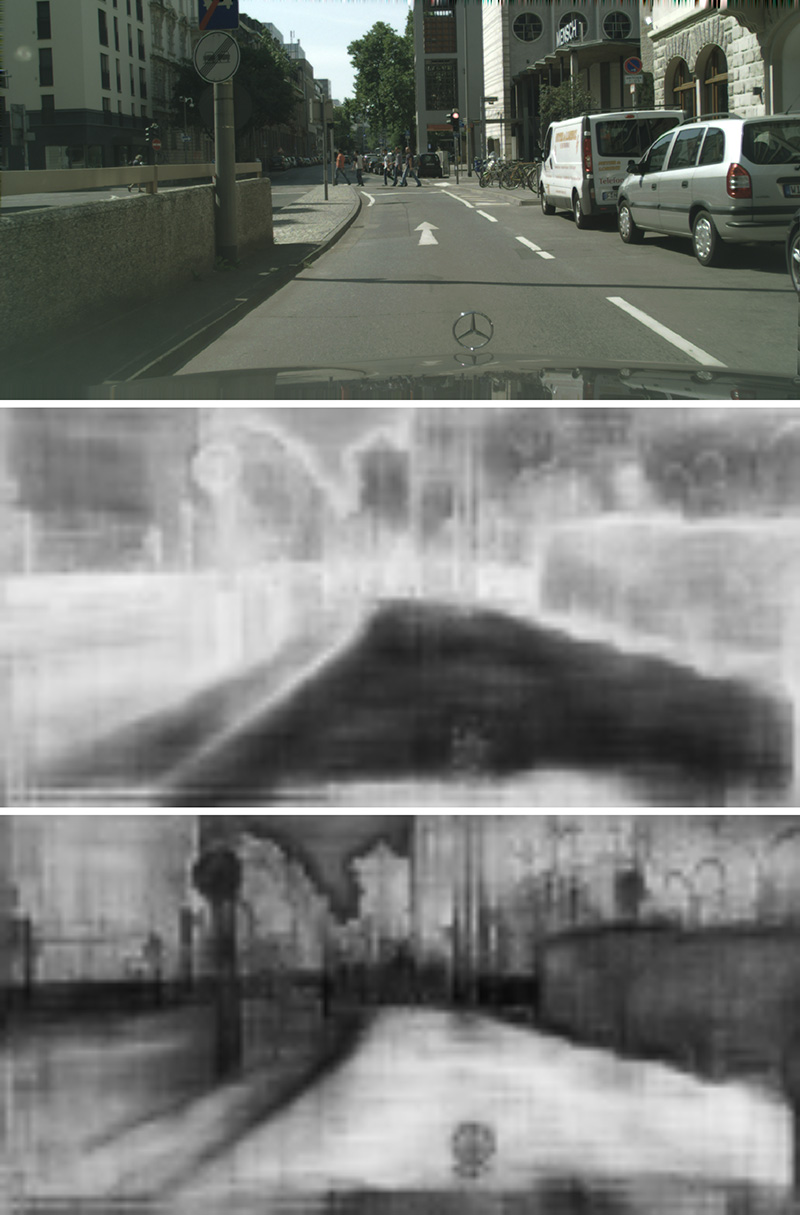}
	\end{subfigure}
	\par\smallskip
	\begin{subfigure}{.33\textwidth}
		\centering
		\includegraphics[width=.99\linewidth]{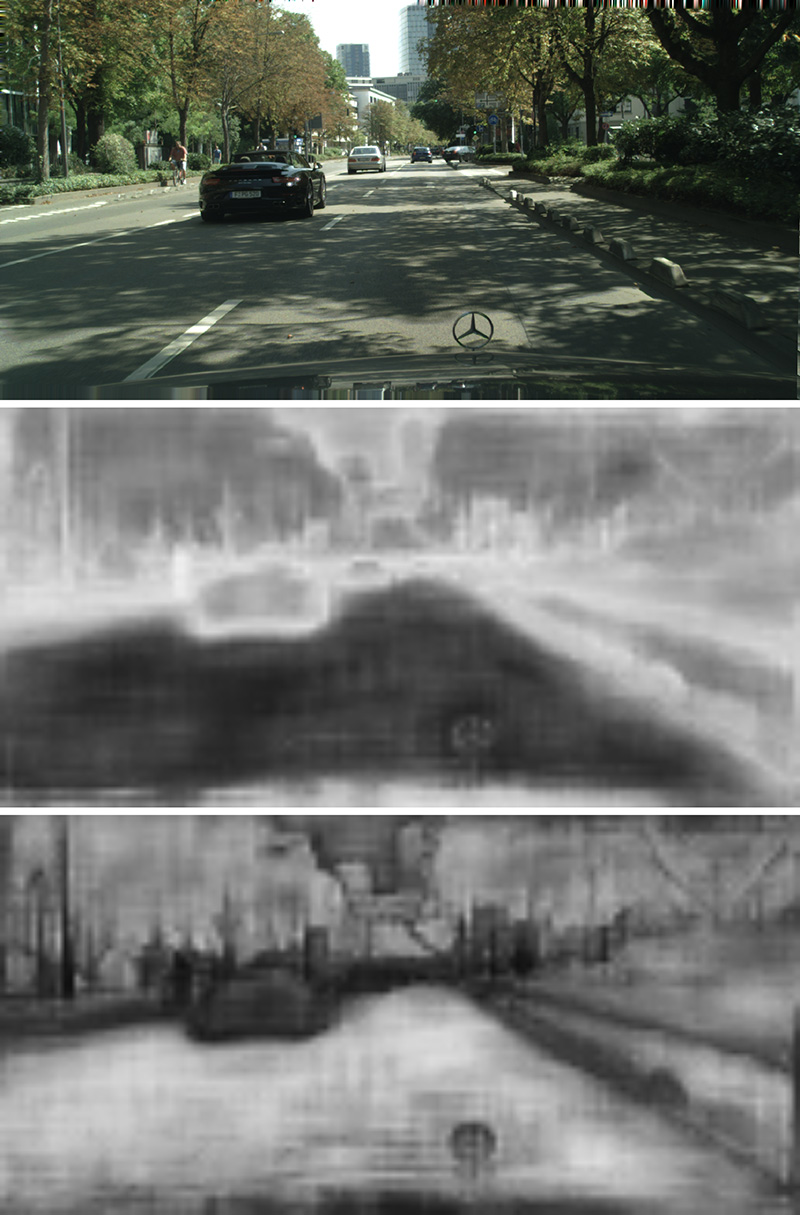}
	\end{subfigure}
	\begin{subfigure}{.33\textwidth}
		\centering
		\includegraphics[width=.99\linewidth]{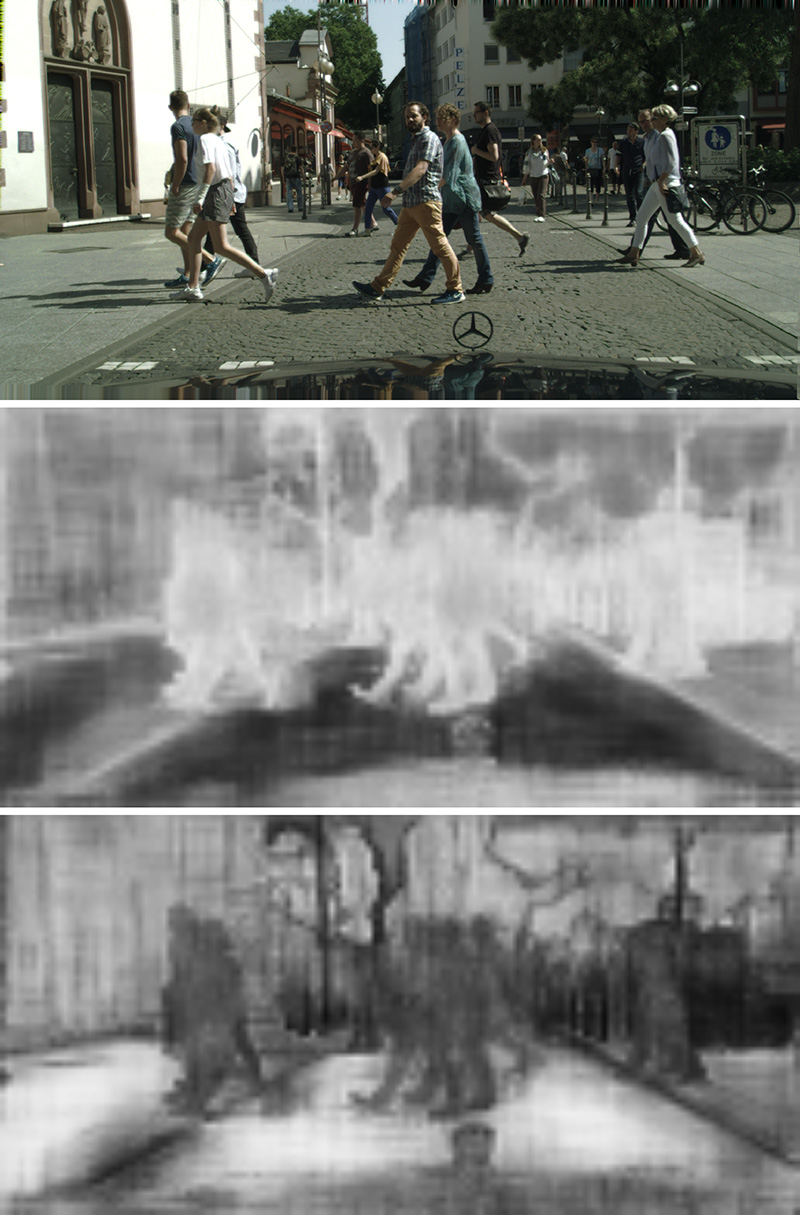}
	\end{subfigure}
	\begin{subfigure}{.33\textwidth}
		\centering
		\includegraphics[width=.99\linewidth]{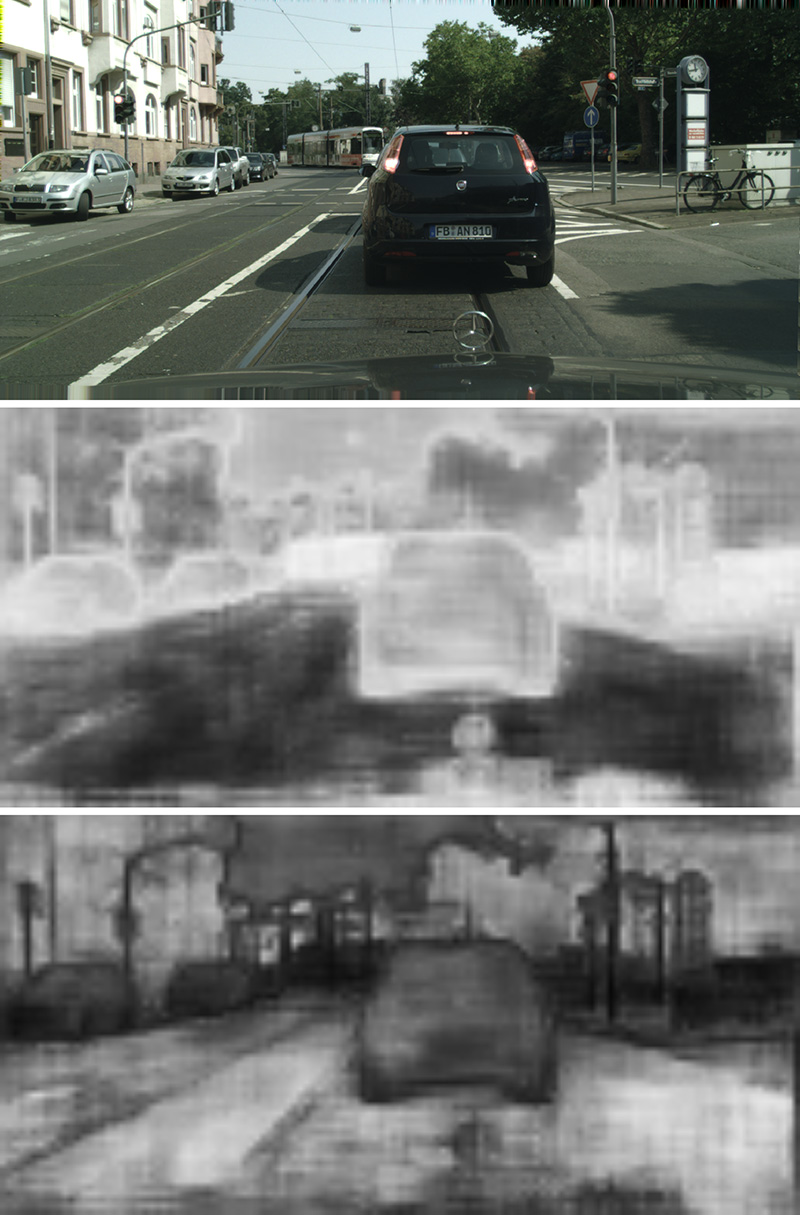}
	\end{subfigure}
	\caption{Visualization of confidence gates values. From top to bottom: input image, confidence gate on appearance network $\sigma_{appr}$, confidence gate on 
	memory network $\sigma_{mem}$. The values are visualized as 0 to black and 1 to white, showing that the network tends to have more confidence on the 
	appearance network for foreground objects, \eg person, car, pole, but more on memory network for background stuff objects, \eg road, sidewalk, terrain.}
	\label{fig:attn}
\end{figure*}

One of the general problems in video modeling is to extract useful spatiotemporal information in order to help with
recognition. Although it is intuitive that videos always contain more information than still images, they also introduce
redundancy and noise. Thus, it is likely that the memory network using spatiotemporal information in some cases performs
not as good as the appearance network using spatial information only. To address this problem, our method uses confidence
gates to control the ensemble of final output. As shown in Fig~\ref{fig:attn}, the network has more confidence on the
appearance network for foreground objects, \eg person, car, pole. On the other hand, for background stuff objects such as
road, sidewalk, terrain, the memory network gets more confidence. Since the values of gates are learned during training,
they show the capability of two networks predicting certain objects, which also indicates the underlying difference of
spatial and spatiotemporal features.

This finding from another perspective explains the improvement of our method gained on certain stuff classes. The gates
serve a role of an attention mechanism that not only lets the network focus on predicting certain classes using more
preferred feature, either spatial or spatiotemporal, but also helps interpret how the deep learning model deals with video
data. One possible explanation is that some stuff classes are more context-dependent than appearance-dependent, and the
spatiotemporal features encodes more context information. We believe there are much more interesting ideas in
spatiotemporal modeling, and expect future research to in this area to be fruitful for both understanding the problem of
perception and for improving the accuracy and robustness of real-world perception systems.

\section{Conclusion}

We show that temporal dynamics information in video of driving scenes contains valuable information for the task of
semantic segmentation. In particular, we find that background classes are more commonly context-dependent and thus
benefit from memory models more than from appearance models. And conversely, foreground objects are more accurately
segmented from appearance information, and do not benefit as much from modeling the object's trajectory in time.  The
\datasetname{} dataset released with this work is used to show the value of temporal dynamics information in this paper
and allows the computer vision community to explore modeling both short-term and long-term context as part of the
driving scene segmentation task.

\section*{Acknowledgments}

This work was in part supported by the Toyota Collaborative Safety Research Center. The views and conclusions being
expressed are those of the authors and do no neccessarily reflect those of Toyota.

{\small
\bibliographystyle{ieee}
\bibliography{lex,segmentation}
}

\end{document}